\documentclass[a4paper]{article}

\usepackage{INTERSPEECH2022}

\title{INTENT CLASSIFICATION USING PRE-TRAINED LANGUAGE AGNOSTIC EMBEDDINGS FOR LOW RESOURCE LANGUAGES 
\\ {\footnotesize Submitted to Interspeech 2022}
}

\name{
    Hemant Yadav\textsuperscript{\rm 1}, Akshat Gupta\textsuperscript{\rm 2}, Sai Krishna Rallabandi\textsuperscript{\rm 3}, Alan W Black\textsuperscript{\rm 3}, Rajiv Ratn Shah\textsuperscript{\rm 1}
}
\address{
    \textsuperscript{\rm 1}IIIT Delhi, India, 
    \textsuperscript{\rm 2}J.P.Morgan AI Research, New York, USA,
    \textsuperscript{\rm 3}Carnegie Mellon University}
    
\email{ 
\{hemantya, rajivratn\}@iiitd.ac.in, \{srallaba, awb\}@andrew.cmu.edu,
akshat.x.gupta@jpmorgan.com
}

\begin{document}

\maketitle
\begin{abstract}
  
Building Spoken Language Understanding (SLU) systems that do not rely on language specific Automatic Speech Recognition (ASR) is an important yet less explored problem in language processing. In this paper, we present a comparative study aimed at employing a pre-trained language agnostic acoustic model to perform SLU in low resource scenarios. Specifically, we use three different embedding settings extracted using Allosaurus, a pre-trained universal phone decoder: (1) Phone-labels (2) Panphone, and (3) Allo embeddings (proposed by us). These embeddings are then used in identifying the spoken intent. We perform experiments across three different languages: English, Sinhala, and Tamil each with different data sizes to simulate high, medium, and low resource scenarios. Our system improves on the state-of-the-art (SOTA) intent classification accuracy by absolute 2.11\% for Sinhala and 7.00\% for Tamil and achieves competitive results in English. Furthermore, we also present a quantitative analysis to show how the performance scales with the number of training examples.

\end{abstract}

\noindent\textbf{Index Terms}: Allosaurus, low resource, dilated CNNs, embeddings, Panphone.

\section{Introduction}
\label{sec:intro}

Spoken language understanding (SLU) systems are fundamental blocks when building interactive technologies for new languages. A typical SLU system consists of an Automatic Speech Recognition (ASR) module followed by a Natural Language Understanding (NLU) module. ASR converts speech to textual transcriptions and the NLU module performs downstream tasks like intent recognition and slot filling from the transcripts obtained. However, building high fidelity ASR systems requires a large amount labelled data which is usually not available for most languages. Language specific ASR system thus forms a bottleneck for creating SLU systems for low-resourced languages. To circumvent this, phonetics based SLU systems have been proposed where the need for language specific ASR is bypassed by typically using a universal phone decoder. This allows creation of language and task specific, word-free, NLU modules that perform intent recognition directly from phonetic transcriptions. 

In this paper, we show that our proposed choice of method \emph{i.e.,} 1-D dilated CNN coupled with Allo embeddings perform competitively with current state-of-the-art (SOTA) SLU systems on English language, and we report new SOTA on Sinhala and Tamil. We work with natural speech datasets in three languages - English, Sinhala and Tamil each with different data sizes to simulate high, medium, and low resource scenarios as shown in Table \ref{table:dataset-statistics}. Our contributions are as follows: (i) We present a 1-D dilated CNN based method coupled with Allo embeddings outperforms the previous approaches that employ phonetic transcriptions (ii) We study the effect of $3$ different embeddings (explained in Section \ref{sec:method}) on the performance of the task \emph{i.e.,} - (a) Phone, (b) Panphone and, (iii) Allo embeddings and (3) We study how the performance scales with the number of training examples. 

\section{Related Works}

Intent recognition has been traditionally performed using textual transcripts generated by ASR systems. Since building ASR technologies is not viable for most languages, recent work has focused on creating such systems using alternate methods. In  \cite{buddhika2018domain}, authors use spectral features of input speech such as MFCCs for intent recognition. NLU modules have also been built for low resourced languages using outputs of an English ASR system, for example, using the softmax outputs of DeepSpeech \cite{hannun2014deep}. DeepSpeech is a character level model where the softmax outputs corresponding to the model vocabulary were used as inputs to the intent classification model \cite{karunanayake2019transfer}. Similarly, softmax outputs of an English phoneme recognition system \cite{lugosch2019speech} have also been used to build intent recognition systems for Sinhala and Tamil \cite{karunanayake2019sinhala}. 

On the other hand,\cite{gupta2020mere}\cite{gupta2021acoustics}\cite{gupta2021intent} proposed to build NLU module using phones extracted from Allosaurus \cite{9054362}. Allosaurus is a universal phone recognizer and therefore language independent. A prototypical naive-bayes intent classifier was built using Allosaurus phonetic transcriptions as inputs in \cite{gupta2020mere}. \cite{gupta2021acoustics} was the first extensive work on using phonetic transcriptions for intent classification on multiple low resourced languages from two language families - Romance and Indic languages. \cite{gupta2021intent} was the first study on building intent recognition systems for natural speech and achieved state-of-the-art results on Tamil. Yet their work was unable to achieve competitive results for languages like English or Sinhala with larger amounts of data. Building on the work of \cite{gupta2021intent}, we propose to use Allosaurus to extract a sequence of dense representation instead of the sequence of discrete phones given an audio file as explained in Section \ref{sec:method}. With this, we are able to achieve close to perfect performance on English and significantly push the SOTA on Sinhala and Tamil.

\section{Methodology}
\label{sec:method}

\begin{figure}
 \centering
 \scalebox{0.3}[0.3]{
    \centering
     \includegraphics{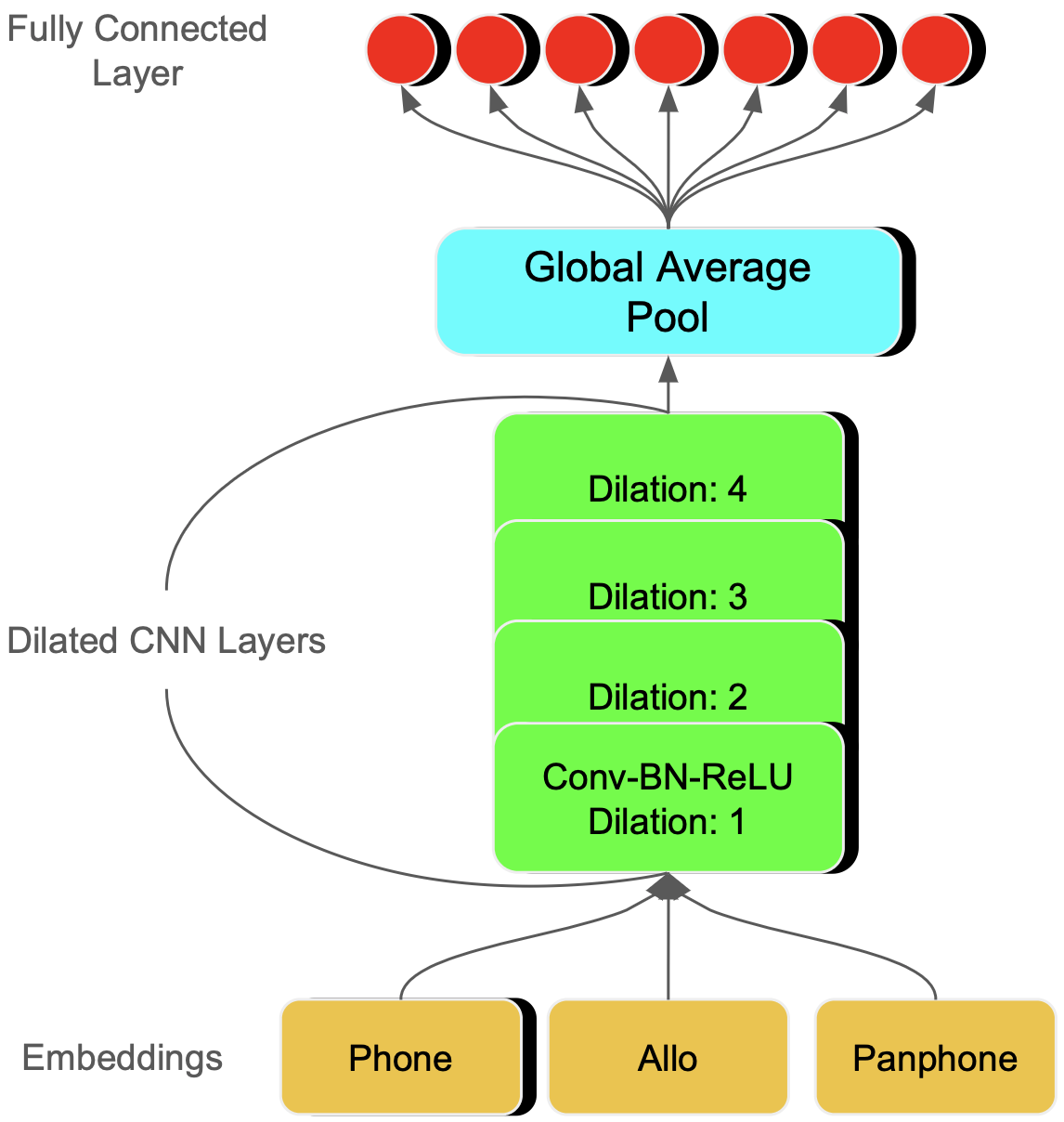}}
     \caption{Our proposed choice of method. The reader must keep in mind that the $3$ different embeddings are used independently \emph{i.e.,} one at a time, to conduct the $3$ different experiments. Any block having a black shadow means the parameters are trainable.}
     \label{fig:model}
\end{figure}

In this section we define our proposed method coupled with the $3$ different input embeddings used for all the experiments. We propose to use an End-2-End 1D dilated convolution neural network (CNN) as shown in Figure \ref{fig:model}. The network consists of 4 CNN-BatchNorm-ReLU-Dropout layers. The CNN filters used are dilated in an increasing linear order from first to fourth layer \emph{i.e.,} 1 dilation in the first layer to 4 dilation in the fourth layer. We apply dilation to increase the overall context. Furthermore we also pad the input to avoid any down-sampling in time dimension. This setup is followed by an average pool and a dropout layer \emph{i.e.,} we map the variable input time steps to a fixed number of time steps, which is $4$. Lastly we add a linear layer to map the output probability distribution over the number of intents.

Let $X = \{x_1, x_2, ..., x_n\}$ be the raw audio signal(input) and $Y = \{y_1, y_2, ..., y_n\}$ be the intent (output). In the first step we map the input to a high-level representation using the Allosaurus tool \footnote{https://github.com/xinjli/allosaurus} \cite{9054362}. The tool is used as a black-box, fixed-weights, and we extract two information from the it (i) The output sequence of phones \cite{moran2014phoible} and (ii) The last layer outputs, before the logit layer, corresponding to each sample $x_i$. Given these two information we define three different embeddings for our proposed method.

\begin{figure*}
 \centering
 \scalebox{0.6}[0.6]{
    \centering
     \includegraphics{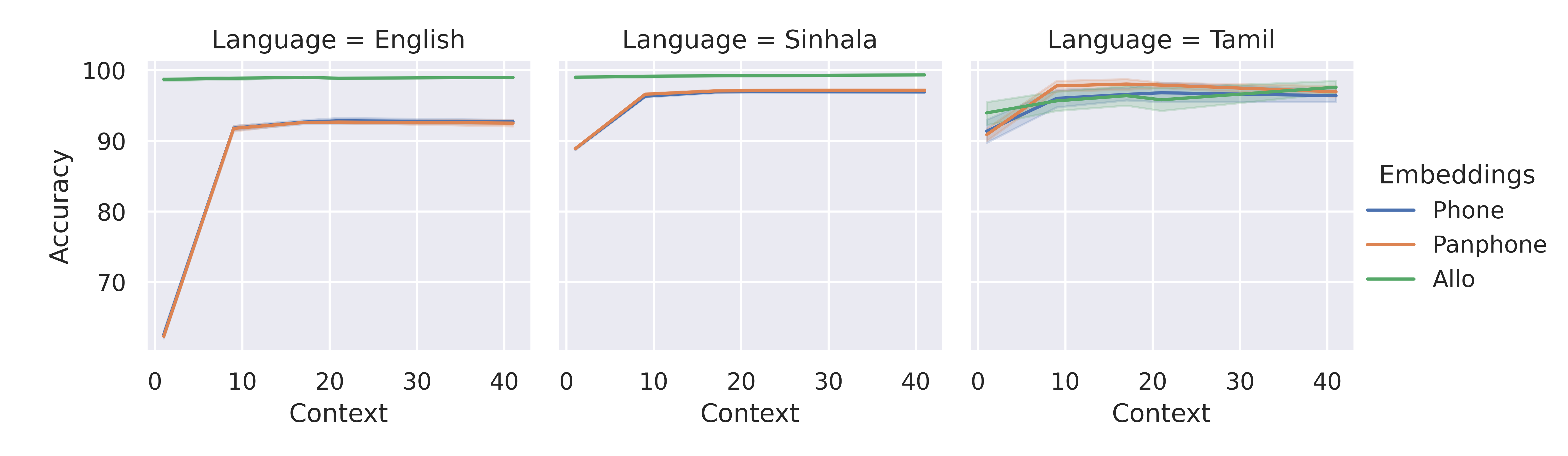}}
     \caption{The plot shows the accuracy vs context-size relation for each of the $3$ different embeddings. Our proposed choice of embedding, Allo, performs the best on all the $3$ languages compared to other two.}
     \label{fig:full_plots}
\end{figure*}

\begin{itemize}
\item Phone ($E_1$): Similar to the previous work\cite{gupta2021intent}, an embedding layer is learnt during the training step such that it maps the individual phones to a 256-dimensional features.

\item Panphone ($E_2$): Instead of learning an embedding layer, we map the individual phone units to a 26-dimensional features similar to the work by \cite{mortensen-etal-2016-panphon}. Therefore the embedding is a 26-dimensional fixed features for each phone.

\item Allo ($E_3$): This is our proposed choice of embedding. The embeddings are language agnostic and our experiments show similar performance on the intent classification task for different languages, Sinhala and Tamil, when the dataset size is comparable as shown in Figure \ref{fig:full_plots} and \ref{fig:scale the training set}. To the best of our knowledge this is a first work to use the pre-trained 640-dimensional last layer of Allosaurus as an embeddings for the intent classification task. We call it Allo embeddings. 

\end{itemize}

From now on we will use the word embedding and input interchangeably \emph{i.e.,} input can be one of the $3$ embeddings explained earlier.

\section{Dataset}
\label{sec:dataset}

\begin{table}
\caption{Dataset statistics for $3$ different languages used in this work \emph{i.e.,} English, Sinhala and,0 Tamil.}
\centering
    \scalebox{1.0}{
    \begin{tabular}{p{1.4cm}|p{1.6cm}|p{1.6cm}|p{1.6cm}}
    \hline
      \textbf{Language} & \textbf{Number of Utterances} & \textbf{Number of Speakers}  & \textbf{Number of Intents} \\
      \hline
     English \cite{lugosch2019speech} & 30,043 & 97 & 31  \\
      Sinhala \cite{buddhika2018domain} & 7624     & 215     & 6 \\
     Tamil  \cite{tamil}  & 400   & 40     & 6 \\
     \hline
        \end{tabular}}

\label{table:dataset-statistics}
\end{table}

\begin{table}
\caption{The table shows 5 different training configurations. Hyphens separate the 4 CNN layers such that 3-5-7-9 means the architecture has a kernel size of 3,5,7,9 for 1,2,3,4 layer number respectively. We also compute the overall context size for an experiment for an easier comparison between the different experiments.}
\label{table:kernel-to-context size mapping}
\centering
    \scalebox{1.0}{
    \begin{tabular}{p{0.4cm}|p{2cm}|p{2cm}|p{2cm}}
    \hline
      \textbf{\%} &\textbf{kernel sizes} & \textbf{Dilation rate} &\textbf{Context size} \\
      \hline
     C1 & 1-1-1-1 & 1-1-1-1 & 1 \\
     C2 &  3-3-3-3 & 1-1-1-1 & 9 \\
     C3 & 3-3-3-3  & 1-2-3-4 & 17 \\
     C4 & 3-5-7-9  & 1-1-1-1 & 21 \\
     C5 & 3-5-7-9  & 1-2-3-4 & 41 \\
     \hline
        \end{tabular}}

\end{table}

In this study, we experiment with $3$ different languages \emph{i.e.,} English, Sinhala and, Tamil with varying training and test sizes and classify them as high, medium and, low resource respectively. The complete statistics are shown in Table \ref{table:dataset-statistics}. For English, we use the  largest freely available Fluent Speech Commands (FSC) dataset \cite{lugosch2019speech}. The dataset has 248 unique sentences spoken by 97 speakers and there is no overlap of speakers between train, valid and, test. Similar to \cite{gupta2021intent}, we use the 31-class intent classification formulation of this dataset.

Sinhala\cite{buddhika2018domain} and Tamil\cite{tamil} datasets are of banking domain collected via crowd-sourcing. Both the datasets have the 6-class intents. Similar to the previous work \cite{gupta2021intent}, we also evaluate our models using 5-fold cross-validation technique \cite{tamil,buddhika2018domain}, since there is no train, development and, test splits provided by the authors.

\begin{figure}
 \centering
 \scalebox{0.5}[0.5]{
    \centering
     \includegraphics{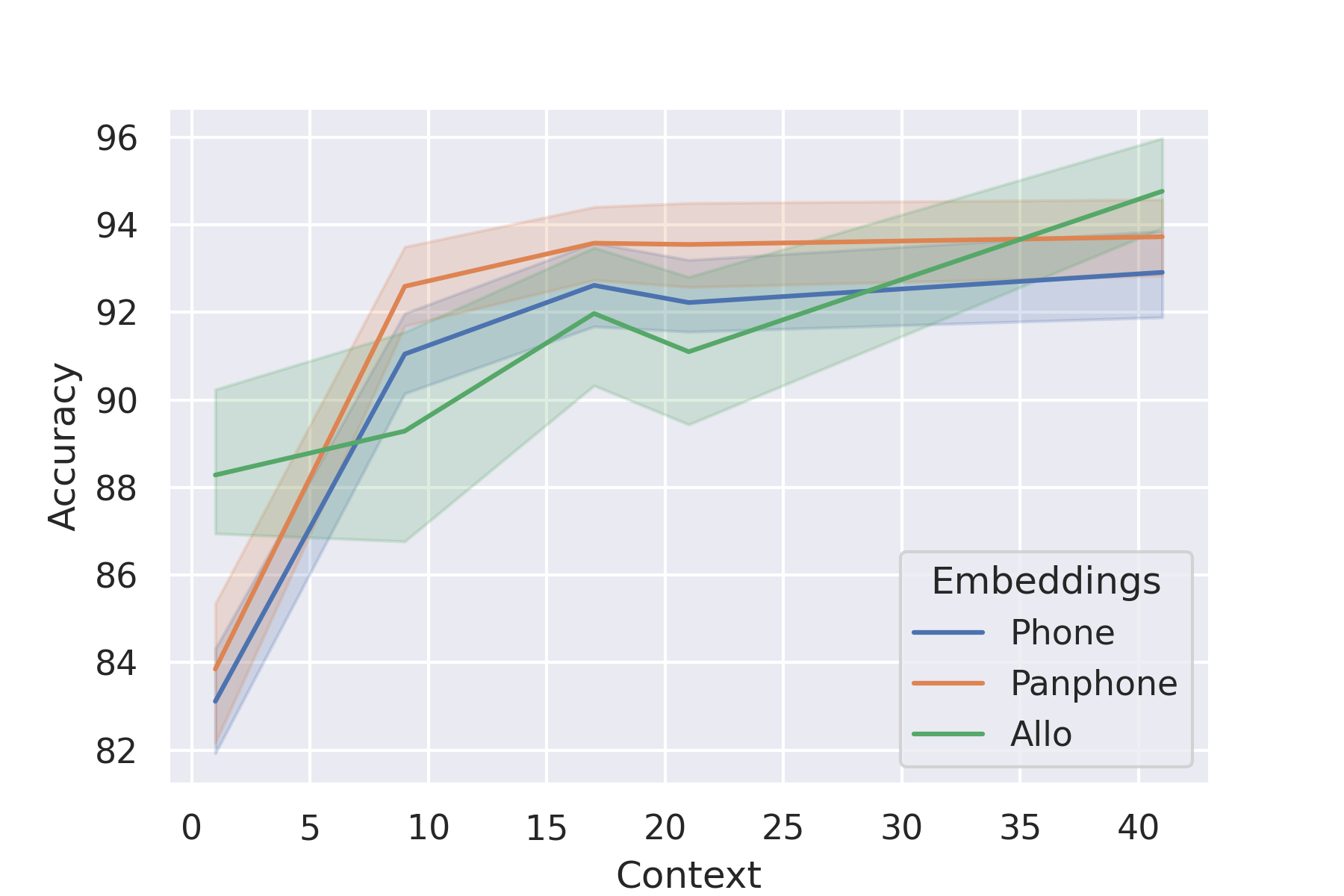}}
     \caption{We plot results on Sinhala language when the training size is similar to Tamil. To study the language agnostic embedding claim made by us. We see a similar trend for all the 3 embeddings, which again are derived from Allosaurus tool which is a language independent phone decoder.}
     \label{fig:sinhala-train-similar}
\end{figure}

\section{Experimental Setup}

We train and evaluate our proposed model on three different languages of varying dataset sizes as explained in Section \ref{sec:dataset}. We fix the number of layers to $4$ and experiment with 5 different configuration of kernel sizes with varying dilation as shown in Table \ref{table:kernel-to-context size mapping}. Here context size refers to the total context the output layer activation has just before the Global pooling layer. Furthermore we map each experiment with its context size since it is more intuitive to reason using context information. Therefore we will use context size to differentiate the experiments instead of kernel size and dilation. We experiment with three different embeddings independently as shown in Figure \ref{fig:model}. Note that out of these three only the Phone embedding ($E_1$) has learnable parameters.

In all experiments, we use L2 weight decay, and dropout as regularization and Adam as an optimizer with 0.0015 learning rate. We decay the learning rate linearly up to 0.000001. The evaluation metrics we report is accuracy. All the reported results are the average of 5 runs with different random seed. Detailed training configurations and the code is available on GitHub\footnote{Will be made available on GitHub upon acceptance.}.

\section{Experimental Results}

In this section we report results to show why CNN is a better architecture choice than LSTM or Transformers in low resource settings \emph{i.e.,}. We compare our proposed method with previous work \cite{gupta2021intent} using Phone embeddings which employed LSTMs and Transformers. Secondly, we compare Phone, Panphone and, Allo embeddings with increasing context size. Lastly we compare the performance of these three embeddings as the number of training example are increased.

Based on our experiment results, in almost all the settings we recommend to use Allo embeddings for the intent classification task. We observe that choosing a bigger context size is a necessity when using our proposed choice of embedding, Allo, in low resource datasets.

\begin{table}
\caption{Accuracy on different architectural and training choices made in the literature when using Phone embeddings. Gupta et al. \cite{gupta2021intent} experiments using LSTMs and Transformer architecture. Our proposed choice of 1-D dilated CNN method shows accuracy gains as the amount of training data decreases. The reader should keep in mind our method is most similar to \cite{gupta2021intent}. Thus a fair comparison would be to the second column.}
\centering
    \scalebox{1.0}{
    \begin{tabular}{p{1.3cm}|p{1.8cm}|p{1.8cm}|p{1.3cm}}
    \hline
      \textbf{Language} & \textbf{End-2-End approaches} & \textbf{Gupta et al. \cite{gupta2021intent}} & \textbf{Our Method} \\
      \hline
    English  & \textbf{99.71\%} \cite{qian2021speech} &  92.77\% & 92.99\% \\
          Sinhala  & \textbf{97.31\%} \cite{karunanayake2019sinhala}     & 96.33\%     & 97.05\% \\
         Tamil   & 81.7\% \cite{karunanayake2019sinhala}  & 91.50\%     &  \textbf{97.25\%}\\
         \hline
        \end{tabular}}
        \label{table:compareprevwork}
\end{table}

\subsection{Comparison With Previous Work Using Phone Embedding}

Similar to \cite{gupta2021intent}, we train our proposed choice of architecture \emph{i.e.,} 1-D dilated CNN, on all the three languages using the phone embeddings (our method). Our method performs better on all the three languages when compared to the similar training settings using LSTM and Transformers by \cite{gupta2021intent}. We also compare our method with End-2End systems in the literature \cite{qian2021speech, karunanayake2019sinhala}. The accuracy gap increases as the dataset size decreases as shown in Table \ref{table:compareprevwork} and we even beat End-2-End approach by a large margin on Tamil. 

Based on these observations, we can now say that CNN is a better choice for intent classification task in low resource nd performs on par in high resource settings. For Tamil we report a new SOTA accuracy and for Sinhala we achieve near SOTA accuracy when using the phone embeddings $E_1$ with our proposed choice of architecture.

\begin{figure*}
 \centering
 \scalebox{0.6}[0.6]{
    \centering
     \includegraphics{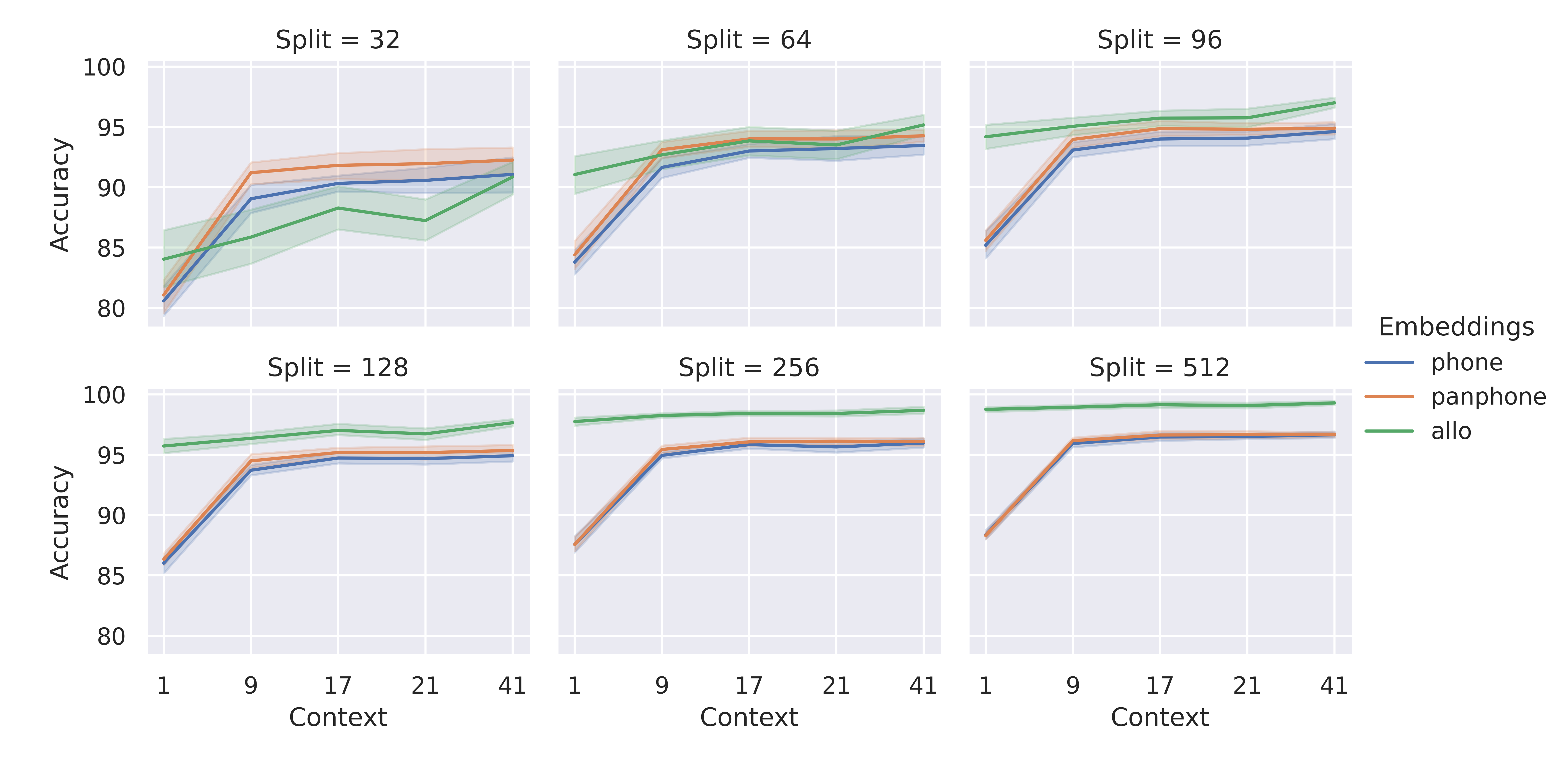}}
     \caption{This plot shows the accuracy vs context size on using $3$ different embeddings as we increase the number for training examples per intent.}
     \label{fig:scale the training set}
\end{figure*}

\subsection{Comparing the Phone, Panphone and, Allo Embeddings}

Our proposed choice of embeddings, Allo, achieves the best accuracy on all the three languages when compared to Phone and Panphone embeddings. 1-D dilated CNN coupled with Allo embeddings achieve a new SOTA on Sinhala and Tamil as shown in Table \ref{table:allo1d} and Figure \ref{fig:full_plots}. When compared to End-2-End system by \cite{qian2021speech} on English language our method performs on par. The reader should keep in mind that additional gains can be seen when fine-tuning the Allosaurus tool in an End-2-End fashion with 1-D dilated CNN. It has to be noted that compared to some of the earlier works \cite{gupta2021intent,karunanayake2019sinhala}, our proposed method works exceptionally well in case of medium and low resource languages \emph{i.e.,} Sinhala and Tamil because of the proposed choice of architecture which is shown to be less prone to overfitting. 

Interestingly we observe that in case of Tamil, a low resource language in our current setup, Allo embeddings does not provide significant gains in accuracy compared to Phone and Panphone. We wanted to test if the cause for this behaviour is the small training dataset or the language itself. Therefore we sample training data from Sinhala of similar size to Tamil and repeat the same experiments. As shown in Figure \ref{fig:sinhala-train-similar}, we observe a similar pattern as before across all the three embeddings \emph{i.e.,} as was seen for Tamil language as shown in Figure \ref{fig:full_plots}. In extremely low resource setting Allo embeddings with bigger context size performs on par to Phone and Panphone embeddings. And therefore it should be the de-facto choice compared to Phone and Panphone embeddings. Based on the similar performance on Tamil and Sinhala, we say that the Allo embeddings are language agnostic as expected and the behavior is dependent on the amount of training dataset used.

Finally, in high and medium resource settings, the Allo embeddings perform similar no matter the context size. This shows that Allo embeddings has contextual information too. In low resource we see the opposite behavior behavior, our hypothesis is that the Allo embedding features are not discriminative enough and a bigger context size somewhat compensates for dataset size. We validate our hypothesis in the next Section.



\subsection{Allo Embedding Performance VS Number Of Training Examples}


Given our previous observations with Tamil, we were interested in the correlation between the number of training examples and the our proposed choice of model \emph{i.e.,} number of training example vs accuracy. Therefore we scale the training dataset size such that $n$*split is the number of training examples, where $n$ is the number of intents. For example in case of Sinhala language we have $n$ equal to 6 and if the value of split is 32, this would give us 192 training examples. We vary the value of split from 32 to 512 such that the number of training examples ranges to 192 to 3072 as shown in Figure \ref{fig:sinhala-train-similar}. We use the model with the highest context size.

We experiment on Sinhala to test the trend of accuracy using different embeddings as we increase the number of training examples. We choose Sinhala language and not English because the majority of Allosaurus training data was English and therefore the results could be biased. As shown in Figure \ref{fig:scale the training set} we observe that the performance of Allo embeddings is proportional to the number of training examples and saturates after a certain point. Furthermore with only 192 training examples (split=32) Allo perform on par with Phone Panphone embeddings given that the model has a higher context size. After the split value of 64 Allo embedding gains significant upper hand in accuracy compared to Phone and Panphone embeddings. These experiments also validate our earlier hypothesis that context size is compensating for the lack of dataset when using Allo embeddings.

\begin{table}
\caption{Comparing the Allo embedding with the other two for each language. Experiments are conducted using the configuration number $5$ as shown in Table \ref{table:dataset-statistics} \emph{i.e.,} the biggest context size.}
\centering
    \scalebox{1.0}{
    \begin{tabular}{p{1.7cm}|p{1.7cm}|p{1.7cm}|p{1.7cm}}
    \hline
      \textbf{Language} & \textbf{Phone} & \textbf{Panphone} & \textbf{Allo} \\
      \hline
    English  & 92.99\%  &  92.96\% & 99.08\% \\
          Sinhala  & 97.05\%     & 97.36\%     & \textbf{99.42\%} \\
         
         Tamil   & 97.25\%  & 97.75\%     &  \textbf{98.50\%}\\
         \hline
        \end{tabular}}
        \label{table:allo1d}
\end{table}

\section{Conclusion and Future work}

In this work we propose language agnostic embedding coupled with 1-D CNN based architecture for the intent classification task which achieves new SOTA accuracy in medium and low resource settings \emph{i.e,} for Sinhala and Tamil language respectively and performs on par on high resource \emph{i.e.,} English. We observe that for Allo embedding to perform on par with Phone and Panphone embedding in low resource settings bigger context size is needed to compensate for the dataset size. Similarly to \cite{gupta2021intent}, our proposed method can also be extended to do slot identification task. For the future work, we would like to explore how to make the Allo embeddings work better in extremely low resource settings \emph{i.e.,} with a smaller context size.

\bibliographystyle{IEEEtran}

\bibliography{main}


\end{document}